\documentclass[conference]{IEEEtran}

\usepackage{amsmath,amssymb,amsfonts}
\usepackage{graphicx}
\usepackage{svg}
\usepackage{textcomp}
\usepackage{xcolor}
\usepackage{url}
\usepackage{booktabs}
\usepackage{multirow}
\usepackage{subfigure}
\usepackage{array}

\usepackage{pifont}  
\usepackage{balance}
\usepackage{textcomp}
\usepackage{comment} 
\usepackage{float}
\usepackage{algorithm}
\usepackage{rotating}
\usepackage{wasysym}
\usepackage{hyperref}
\usepackage{subcaption}
\usepackage{afterpage}
\usepackage{enumitem}
\usepackage{stfloats}
\usepackage{comment} 
\usepackage{rotating}
\usepackage{placeins}
\def\BibTeX{{\rm B\kern-.05em{\sc i\kern-.025em b}\kern-.08em
    T\kern-.1667em\lower.7ex\hbox{E}\kern-.125emX}}
    
\begin{document}

\title{Do No Harm? Hallucination and Actor-Level Abuse in Web-Deployed Medical Large Language Models}


\author{
\IEEEauthorblockN{
Sunday Oyinlola Ogundoyin\IEEEauthorrefmark{1},
Muhammad Ikram\IEEEauthorrefmark{1},
Rahat Masood\IEEEauthorrefmark{2}
}

\IEEEauthorblockA{\IEEEauthorrefmark{1}
Macquarie University, Sydney, Australia\\
\{sunday.ogundoyin, muhammad.ikram\}@mq.edu.au
}

\IEEEauthorblockA{\IEEEauthorrefmark{2}
The University of New South Wales, Sydney, Australia\\
rahat.masood@unsw.edu.au
}
}

\maketitle

\begin{abstract}
Medical large language models (LLMs), including custom medical GPTs (MedGPTs) and open-source models, are increasingly deployed on web platforms to provide clinical guidance. However, they pose risks of hallucination, policy noncompliance, and unsafe design. We conduct a large-scale assessment of 6,233 MedGPTs, evaluating a stratified sample of 1,500, together with 10 open-source LLMs.  We introduce two frameworks: MedGPT-HEval for hallucination detection and an LLM-based pipeline for assessing policy violations and developer intent. Our results show that 25--30\% of MedGPTs exhibit low factual accuracy, with bottom- and middle-tier models at highest risk; 33.6--54.3\% violate operational thresholds, and 57.06\% of Action-enabled models lack adequate privacy disclosures. Compared with open-source models, MedGPTs achieve higher factual accuracy and semantic alignment, though open-source models are more stable. These results reveal systemic gaps in hallucination and compliance, highlighting the need for multi-metric evaluation and stronger safeguards. We release HAA-MedGPT, a structured dataset that supports future research on the safety of web-facing medical LLMs.
\end{abstract}


\section{Introduction}\label{sec:intro}

Large language models (LLMs) are increasingly deployed as user-centric applications on commercial platforms, where users interact with configurable agents, such as those in the OpenAI store. Among these, custom-built medical GPTs (MedGPTs) provide diagnostic suggestions, health advice, treatment explanations, and symptom checks~\cite{ALSABAWI2025}. The GPT Store~~\cite{OpenAIStore} hosts thousands of such models, marketed for consultation, triage, and education~\cite{GPTRACKER}. Despite automated and human reviews, unsafe or abusive MedGPTs continue to appear~\cite{Ogundoyin-LLM}, underscoring persistent safety risks in platform-level deployments. In parallel, open-source medical LLMs such as Galactica~\cite{galactica}, PMC-LLaMA~\cite{pmc-llama}, and MedAlpaca~\cite{medalpaca} prioritize transparency and flexibility. However, they often fall short on diagnostic performance~\cite{gumilar20244019, AHMED2025, PAGANO20259}, showing a trade-off between accessibility and clinical accuracy. These ecosystems illustrate a key tension in medical LLM deployment: platform-hosted MedGPTs deliver high factual accuracy but carry safety risks, while open-source models offer transparency but may lack reliability. This motivates systematic evaluation of factual correctness, consistency, and safety in medical LLMs.

MedGPTs expose users to two critical but often overlooked risks. The first is clinical hallucination -- confident but false or fabricated medical information~\cite{Huang_2025, Bang2025}, such as unsafe treatments or incorrect drug advice. The second is design-level abuse, where GPT creators violate OpenAI's privacy policies~\cite{rodriguez2025safer} or manipulate users through deceptive names, descriptions, or by using tools that circumvent safeguards. These risks are amplified by platform trust indicators (such as ratings and reviews) that falsely legitimize unsafe GPTs. Harm arises from both outputs and model configuration. Developers craft names, descriptions, and conversation starters that convey trust or authority and may enable tools such as web browsing~\cite{hou_security_2024}, while linking to vague or broken privacy policies. These surface-level cues shape users' trust and introduce risk before they interact with models. Unlike general domains, where users can spot LLM errors, patients often lack the medical knowledge to evaluate LLM-generated advice~\cite{kim2025medicalhallucinationsfoundationmodels}.

Recent studies show that clinical hallucinations persist in LLMs despite domain tuning and expert evaluation. Kim et al.~\cite{kim2025medicalhallucinationsfoundationmodels} proposed a hallucination taxonomy and found that Med-PaLM~\cite{Med-PaLM} and GPT-4~\cite{GPT-4} still produce misleading outputs, even with retrieval-augmented prompting. Asgari et al.~\cite{Asgari2025} similarly observed factually unsupported but fluent medical claims from GPT-4, raising concerns about trust and verifiability. Broader GPT ecosystem studies have focused on jailbreaks, usage patterns, and configuration extraction~\cite{Ogundoyin-LLM, hou_security_2024, zhao2024gptswindowshoppinganalysis}, but treat these issues as isolated failures. No prior work has systematically analyzed clinical hallucination or actor-level abuse in real-world MedGPTs on the web GPT marketplaces~\cite{gptsapp.io, GPTStore.AI, OpenAIStore}. To fill this gap, we introduce HAA-MedGPT, the first large-scale dataset and evaluation framework for detecting hallucinations and intent-level misuse in MedGPTs. Our approach combines multi-metric scoring with policy-aligned analysis to uncover structural risks in model outputs and how they are built, presented, and deployed.

We crawled metadata from 6,233 MedGPTs on the OpenAI Store. We selected a stratified sample of 1,500 models to balance platform coverage while respecting OpenAI's query limits and maintaining responsible, non-disruptive interaction with the GPT Store. Models were grouped by conversation count into three tiers: Top 1000, Middle 250 (random), and Bottom 250. Each GPT was evaluated using standardized clinical prompts and a multi-metric scoring framework. In parallel, we included 10 open-source medical LLMs -- such as Galactica~\cite{galactica}, PMC-LLaMA~\cite{pmc-llama}, and MedAlpaca~\cite{medalpaca} -- in the analysis to compare platform-hosted models with transparent, community-driven alternatives. We also assessed actor-level risk using a rubric-based analysis of names, descriptions, conversation starters, and policy statements, enabling a detailed evaluation of platform-hosted and open-source medical LLMs.

Unlike prior studies that examine hallucination in base LLMs~\cite{Asgari2025} or misuse in general GPT applications~\cite{GPTRACKER}, our work is the first to empirically analyze deployed, user-facing, healthcare GPTs at web scale, capturing both content-level hallucinations and developer-driven misuse signals. 
We ask the following four core questions.

\begin{itemize}[left=0pt]
         \item \textbf{{$\mathbf{RQ1}$}:} How do rates of clinical hallucination in MedGPTs vary across popularity tiers on the OpenAI Store?
    \item \textbf{{$\mathbf{RQ2}$}:} Are users able to discern hallucinated content in MedGPTs' outputs?
      \item \textbf{{$\mathbf{RQ3}$}:} How do developer-defined design choices enable abusive behaviors or weaken privacy safeguards in MedGPTs?
      \item \textbf{{$\mathbf{RQ4}$}:} How do clinical hallucinations differ between MedGPTs and open-source medical LLMs?
\end{itemize}

To assess RQ1, we introduce MedGPT-HEval  (\S\ref{subsec:results_hallucination}), a structured multi-metric framework for evaluating clinical hallucination in MedGPTs. We queried each model using a clinical scenario extracted from the MedQA benchmark~\cite{MedQA} and repeated the same question five times to obtain multiple responses. This approach captures variability in the model's outputs and allows for a more robust assessment of hallucination. The responses were then evaluated using four metrics: G-Eval~\cite{G-Eval}, BARTScore~\cite{BARTScore}, semantic entropy~\cite{Semantic_Entropy}, and cosine similarity~\cite{Cosine_similarity}, capturing both factual alignment and consistency. We found that 25--30\% of MedGPTs across top, middle, and bottom tiers score below 0.8 in G-Eval, with only 37.27\% achieving BARTScore $\ge -3.5$ and 41.07\% reaching cosine similarity $\ge 0.4$. At the same time, 59.87\% have semantic entropy less than 2, indicating moderate response stability. Bottom- and middle-tier MedGPTs pose the highest hallucination risk and weakest contextual alignment. This shows that model popularity is not a reliable indicator of factual accuracy or safety in web-deployed medical GPTs.

To evaluate RQ2, we analyzed conversation volume, star ratings, and review sentiment for the Top 1000 MedGPTs (\S\ref{subsec:results_users_awareness}). Analysis shows that user engagement does not reflect awareness of clinical hallucinations, with near-zero correlations for conversation counts (G-Eval: -0.0347; BART: -0.0196; entropy: 0.0057; cosine: -0.0318) and reviews (-0.0449 to 0.0732). In contrast, reviews correlate strongly with usage ($r = 0.9999$ for positive reviews, $r = 0.9656$ for negative reviews). This indicates that user feedback reflects activity rather than accuracy, revealing a gap between perceived trust and actual reliability in web-deployed MedGPTs.

To assess RQ3, we applied an automated scoring pipeline based on OpenAI's operationalized usage policies. We used K-means clustering to determine a cut-off to classify MedGPTs as compliant or noncompliant (\S\ref{subsec:results_actor-defined}). Misuse is common among MedGPTs: 54.3\% of Top 1000, 48.0\% of Middle 250, and 33.6\% of Bottom 250 models exceed the 0.45 risk threshold, and two- or three-violation cases affect 33--64\% across popularity tiers. Privacy and compliance gaps are also prevalent among 170 MedGPTs with Actions enabled (\S\ref{subsec:results_actions_capability}): only 42.94\% had accessible privacy policies (57.06\% lacked documentation) and nearly 70\% of extracted policies scored below the threshold, exposing users to unsafe data practices and regulatory violations.

To assess RQ4, we investigate clinical hallucination in the open-source medical LLMs and compare the results with the MedGPTs (\S\ref{subsec: Hallucination_in_open-ssource LLMs}). Open-source medical LLMs show a trade-off between accuracy and consistency: Galactica (G-Eval: 0.6480) and Aloe-Alpha (G-Eval: 0.5948) have the highest factual accuracy, while MedAlpaca (0.4863), Apollo (0.4863), and MentalHealthChatbot (0.4354) achieve stronger semantic alignment, and BioMistral (2.3978) is the most variable. Compared with MedGPTs, which exhibit higher G-Eval (0.9238) and cosine similarity (0.4054) but greater entropy (1.9272), these results indicate that MedGPTs provide superior factuality and semantic coherence. In contrast, open-source models remain more stable and predictable, underscoring the need for multi-metric evaluation of hallucination risk.

\noindent\textbf{Our Contributions.} Specifically, the contributions of this paper are as follows.
\begin{itemize}[left=0pt]

\item \textbf{Web-scale audit of MedGPTs}.  We design and deploy the first scalable measurement pipeline for auditing web-deployed medical LLM ecosystems, combining automated discovery, interaction-based inference probing, metadata extraction, and policy-compliance analysis across 6,233 deployed agents.

\item \textbf{Dual-layer safety analysis}.  
We introduce MedGPT-HEval for clinical hallucination detection and a complementary actor-level misuse evaluator, jointly exposing risks overlooked in prior work.

\item \textbf{Evidence of structural governance failures}.  
We show that platform trust indicators (ratings, reviews, conversation counts) do not correlate with safety, and that 49.8\% of models violate operational policy.

\item \textbf{Privacy-risk quantification}.  
We provide the first empirical assessment of privacy policy alignment for Action-enabled GPTs, showing that 57.06\% lack functional policy disclosures.

\item  \textbf{Public dataset and tools}.  
We release HAA-MedGPT\footnote{\url{https://anonymous.4open.science/r/HAA-MedGPT-2E78}}--the first large-scale dataset of 6,233 custom web-deployed MedGPTs from the OpenAI Store, enabling future Web-safety, platform-governance, and public-health research.
\end{itemize}

\section{Related Work}
In this section, we discuss previous studies in the literature that are related to our work.
\vspace{-0.3cm}
\subsection{LLM Deployment in Healthcare}
Recent studies have examined LLMs in clinical settings, focusing primarily on foundation models or isolated tasks in controlled environments. Ahmed et al.~\cite{Ahmed2025cardio} and Shekhar et al.~\cite{SHEKHAR2025} explored ChatGPT's potential in cardiovascular care and ambulance triage, respectively, but their works remain conceptual. Benchmarking efforts by Gumilar et al.~\cite{GUMILAR2024} and Pagano et al.~\cite{PAGANO20259} evaluated LLMs such as GPT-4, GPT-4o, LLaMA-3.1, and Copilot on oncology and orthopedics, focusing solely on accuracy. Extensive reviews~\cite{HE2025102963,MENG2024109713} raised concerns around hallucination and ethical opacity, but lacked empirical deployment analysis. Governance frameworks such as Health-LLM~\cite{yu2025healthllm} and Polaris~\cite{mukherjee2024polariss} offer safety-aware designs but operate in tightly controlled or simulated environments. Apparently, none of these efforts evaluates the real-world risks of custom MedGPTs. In contrast, our work systematically evaluates deployed MedGPTs for hallucination and actor-driven abuse.

\subsection{Clinical Hallucinations in LLMs}
Several studies examine hallucinations in medical LLMs using expert benchmarks, taxonomies, and error analyses. Kim et al.~\cite{Kim2025hallu} proposed a typology (diagnostic, factual, outdated) across models (e.g., GPT-4o, PMC-LLaMA~\cite{pmc-llama}, MedAlpaca-13B~\cite{medalpaca}). Asgari et al.~\cite{Asgari2025} and Vishwanarth et al.~\cite{vishwanath2024hallu} built clinician-in-the-loop audit tools (e.g., CREOLA) for summarization. Qin et al.~\cite{qin2024hallu} used entropy-based dialogue modeling to address misinformation but did not target LLM-originated hallucinations or unsafe design. Zhu et al.~\cite{Zhu2025hallu} offered a unified taxonomy linking hallucinations to data, training, and inference, but not to deployed GPTs or developer misuse. Agarwal et al.~\cite{agarwal2025hallu}'s MEDHALU output faithfulness over author intent. Collectively, these studies frame hallucination as a content problem, overlooking structural risks tied to GPT authorship and deployment. Our work fills this gap by analyzing both output-level hallucinations and actor-driven factors in deployed MedGPTs.

\subsection{Actor-Level Abuse and Intent in GPT Deployments}

Recent large-scale audits probe misuse in public LLM apps but rarely target healthcare. Zhang et al.~\cite{zhang2024lookgptappslandscape} analyzed 10,000 GPT apps for prompt and configuration leaks; Hou et al.~\cite{hou_security_2024} surveyed 786,000 apps, revealing data over-collection and deceptive logic for general-purpose abuse. Shen et al.~\cite{GPTRACKER}’s GPTracker flagged over 2,000 Store violations without assessing hallucination or health-specific harm. Rodríguez et al.~\cite{rodriguez2025safer} proposed red-teaming compliance but focused on non-medical, single-turn prompts. These studies demonstrate scalable assessment methods but overlook how developer choices create clinical risk. Our work pairs hallucination detection with design-level analysis to expose how author decisions drive real-world clinical failures.

In summary, prior work either examines hallucinations in base models or analyzes generic misuse, without addressing risks in custom MedGPTs. No existing study evaluates how developer choices drive clinical hallucination or abuse. To fill this gap, we present HAA-MedGPT, the first large-scale dataset and framework for detecting \textbf{H}allucination and \textbf{A}ctor-level \textbf{A}buse in OpenAI-deployed \textbf{Med}ical \textbf{GPT}s. In the next section, we detail our data collection and analysis methods.

\section{HAA--MedGPT: Dataset for Hallucination and Actor-level Abuse Detection in MedGPTs}
In this section, before discussing the data collection and analysis methods, we briefly describe clinical hallucination and actor-abuse vectors, and conclude by highlighting the ethical considerations.

\subsection{Clinical Hallucination in Medical LLMs}\label{subsec:hallu}

Clinical hallucinations are outputs from medical LLMs that are fluent but factually incorrect, incomplete, or irrelevant~\cite{Kim2025hallu,chaudhury_hallucination}. These errors fall into two types: factuality (false, outdated, or unverifiable content) and faithfulness (contradictions or deviations from the user prompt)~\cite{schmidgall2024addressingcognitivebiasmedical}. Unlike obvious falsehoods in general LLMs, clinical hallucinations use domain-specific language and coherent logic, making inaccuracies appear plausible~\cite{Huang_2025}. This increases their danger, as such errors can misguide diagnoses, treatments, or prescriptions~\cite{Kim2025hallu}. Prior work has developed taxonomies, benchmarks~\cite{Kim2025hallu,agarwal2025hallu}, and tools like CREOLA~\cite{Asgari2025} to assess impact. However, detection remains difficult, especially for non-experts, since subtle flaws often pass unnoticed~\cite{Asgari2025}. When accepted uncritically, these outputs erode trust and pose real clinical risks~\cite{pal2023hallu}. As medical LLMs gain real-world use, addressing hallucinations is not just a technical task but a clinical safety imperative.

\subsection{Actor-level Abuse in MedGPTs}\label{subsec:abuse}

Actor-level abuse refers to intentional design choices by GPT creators that predispose MedGPTs to unsafe or unethical use before any user interaction. Unlike hallucinations, which emerge during generation, these risks are embedded in metadata--names, descriptions, conversation starters, and privacy policies--which shape user expectations and may falsely signal clinical authority. Prior work has shown that such configurations can encode malicious intent and bypass safeguards~\cite{GPTRACKER, hou_security_2024, rodriguez2025safer}. In healthcare, this risk is magnified. For example, models named ``Instant Diagnosis AI'' or ``Cancer Risk Estimator'' may promote unverified treatments or simulate diagnostic authority, encouraging unsafe reliance without disclaimers or expert review. The threat grows with Actions-enabled GPTs, which connect to external APIs and process real-time data. As public-facing MedGPTs proliferate, detecting and mitigating actor-level abuse is essential for safety, trust, and responsible deployment.

\subsection{Open-source Medical LLMs}
Recently, there has been significant interest in developing and deploying LLMs tailored to medical domains. These models are promising in diagnostic and clinical decision-making reliability \cite{zhou2024surveyllms}. In this work, we considered 10 open-source medical LLMs, which differ in training data, fine-tuning methods, and evaluation benchmarks, as shown in Table~\ref{tab:open-source model_overview}. These models are selected due to their popularity, recency, and support for text generation. 

\begin{table*}[h!]
    \centering
   \footnotesize
    \caption{Overview of Open-source medical LLMs~\cite{HE2025-Survey-MedLLMs, Liu2024-Survey-MedLLMs}. Here, PT -- Pre-training, STF -- Supervised Fine-tuning, DPO -- Direct Preference Optimization, RLHF -- Reinforcement Learning with Human Feedback}
    \vspace{-0.2cm}
    \tabcolsep = 0.1cm
    \scalebox{0.90} {\begin{tabular}{llllll}
    \toprule
        \textbf{Model} & \textbf{Method} & \textbf{Training data} & \textbf{Evaluation datasets} & \textbf{Released date} & \textbf{Size (Billion)} \\
        \midrule
        Galactica~\cite{galactica} & PT + SFT & DNA, AA Sequence & MedMCQA, PubMedQA, Medical Genetics & 11/22 & 120 \\
        ChatDoctor~\cite{chatdoctor} & SFT & Patient-doctor dialogues & iCliniq & 03/23 & 7 \\
        MedAlpaca~\cite{medalpaca} & SFT & Medical QA and dialogues & USMLE, Medical Meadow & 04/23 & 13 \\
        PMC-LLaMA~\cite{pmc-llama} & SFT & Biomedical academic papers & PubMedQA, MedMCQA, USMLE & 04/23 & 7 \\
        JMLR~\cite{jmlr} & SFT & MIMIC-IV, Medical textbooks, PubMed & USMLE, Amboss, MedMCQA, MMLU-Medical & 02/24 & 13 \\
        BioMistral~\cite{bioMistral} & PT + SFT & PubMed central & MMLU, USMLE, MedMCQA, PubMwdQA & 02/24 & 7 \\
        Apollo~\cite{apollo} & PT + SFT & Books, clinical guidelines, Encyclopedia & XMedBench & 03/24 & 7 \\
        Aloe-Alpha~\cite{aloe-Aplha} & PT + SFT + DPO & MedQA, CoT, Synthetic data & MultiMedQA, MedMCQA, USMLE, PubMedQA, etc. & 05/24 & 8 \\
        MentalHealthChatbot~\cite{mentalHealthChatbot} & SFT & Specialized mental health dataset & -- & 09/24 & 7 \\
        Zhongjing~\cite{Zhongjing} & PT + SFT + RLHF & Medical books, ChatMed, Medical Wiki & CMtMedQA, Huatuo-26M & 04/24 & 13 \\
        \bottomrule
    \end{tabular}}
    \label{tab:open-source model_overview}
\end{table*}

\subsection{Data Collection Pipeline}\label{subsec:data_collection}
To assess the landscape of MedGPTs, we developed a scalable extractor to overcome the visibility limitations of the OpenAI GPT Store, which provides access only to a curated subset. Since GPTs are not fully indexed, our extractor used OpenAI's public search API to collect metadata at scale. We developed a domain-specific taxonomy of 171 medical keywords spanning specialties, roles, conditions, tasks, and general terminology. This taxonomy, generated via a Python-based deduplication pipeline, ensures broad and consistent retrieval of medically relevant GPTs. Using a Selenium-based crawler, we automated searches for each keyword. We extracted metadata fields, including model name, description, author, category, conversation starters, reviews, and capabilities (e.g., browsing, code interpreter, Actions). The pipeline ran from January 20--22, 2026, yielding 9,245 unique GPTs. This dataset reinforces our large-scale analysis of hallucination and actor-level abuse in deployed MedGPTs.

\subsection{Data Analysis}

\subsubsection{Metadata Curation}
To ensure reliability and domain specificity, we manually reviewed all 9,245 collected GPTs based on their names, descriptions, and conversation starters, excluding 3,012 irrelevant entries. This yielded a final set of 6,233 ($\approx$67.42\%) MedGPTs relevant to healthcare contexts. We then cleaned the metadata by removing formatting artifacts (e.g., extraneous ``+'' symbols) and standardizing numeric shorthand (e.g., ``10K'' to 10000, ``1M'' to 1000000). The resulting corpus provides a structured, machine-readable record of MedGPTs, including creator IDs, usage metrics, capabilities, and interface elements. This curated dataset underpins our analyses of hallucination, actor-level misuse, and policy compliance across deployed MedGPTs in the OpenAI Store.

\subsubsection{Overall Distribution}
As shown in Table~\ref{tab:gpt_distribution}, usage and feedback data for MedGPTs in the OpenAI Store reveal major disparities in visibility and oversight. The vast majority (85.58\%, 5,334 GPTs) have under 100 conversations, and 90.81\% (5,660 GPTs) lack any user ratings, suggesting minimal public scrutiny. Engagement is highly skewed: only 15 GPTs have more than 100,000 interactions, and just 3 have more than 5,000 reviews. Even among the reviewed GPTs, quality indicators are weak: only 571 (9.16\%) average above 3.0 stars, raising concerns about reliability. Based on review and conversation counts, we estimate total interactions range from 148,761 to 11.40 million. This imbalance reveals a systemic blind spot, where thousands of MedGPTs remain untested, unvetted, and potentially unsafe. In clinical contexts, where inaccuracies can harm patients, the lack of community feedback makes passive moderation inadequate. These findings underscore the urgent need for proactive safety and vulnerability assessments tailored to MedGPTs.

\subsubsection{Authorship}
\begin{figure}[ht!]
    \centering
    \vspace{-0.2cm}
    \includegraphics[width=0.75\linewidth]{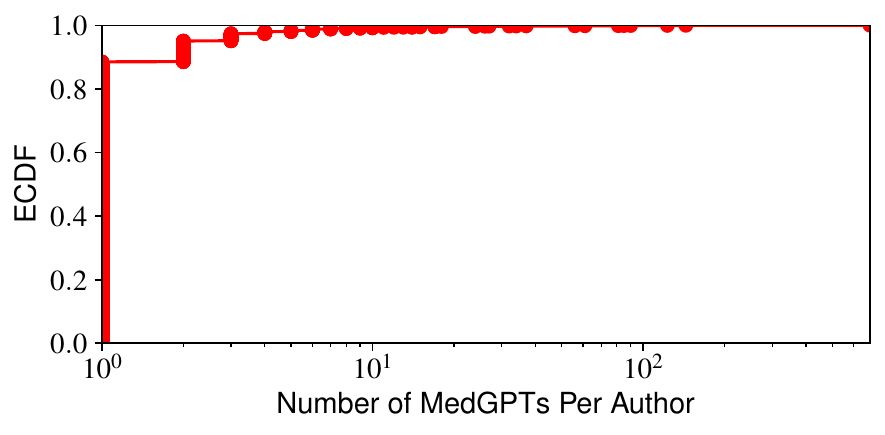}
     \vspace{-0.2cm}
    \caption{ECDF of number of MedGPTs per author/developer.}
    \label{fig:author_cdf}
    \vspace{-0.3cm}
\end{figure}

\begin{figure}[ht!]
    \centering
    \vspace{-0.4cm}
    \includegraphics[width=0.85\linewidth]{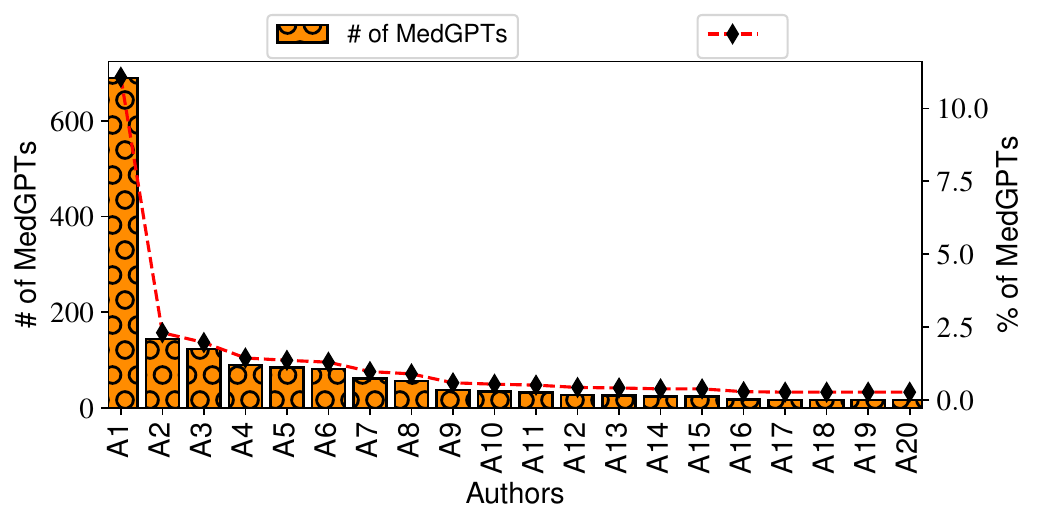}
     \vspace{-0.2cm}
    \caption{Top 20 authors of MedGPTs.}
    \label{fig:top_20_authors}
\end{figure}

Fig.~\ref{fig:author_cdf} shows that medical GPT authorship in the OpenAI Store is highly unbalanced. Over half of developers (53.68\%, or 3,346) have published only one GPT, indicating many are isolated or experimental efforts. In contrast, a small number of creators dominate the ecosystem -- one author alone published 690 GPTs (11.07\%), and the top 20 account for 1,620 GPTs (26\% of the total) (Fig.~\ref{fig:top_20_authors}). This concentration raises safety and governance concerns. While some high-volume contributors may aim to expand content, the platform provides no visibility into their identity, expertise, or quality control. Without transparency or verification, mass-produced GPTs (especially in healthcare) risk spreading low-quality or misleading advice unchecked.

\begin{table}[!t]
    \centering
    \small
\caption{Distribution of MedGPTs metadata.}
\vspace{-0.2cm}
    \label{tab:gpt_distribution}
   \tabcolsep=0.12cm
\renewcommand{\arraystretch}{1.2} 
\resizebox{0.48\textwidth}{!}{\begin{tabular}{llllll}
    \toprule
        \hline
        \multicolumn{2}{c}{\textbf{Conversation counts}} & \multicolumn{2}{c}{\textbf{Average ratings}} & \multicolumn{2}{c}{\textbf{Total reviews}} \\ \hline
        \textbf{Values} & \textbf{\# GPTs} & \textbf{Values} & \textbf{\# GPTs} & \textbf{Values} & \textbf{\# GPTs} \\ \hline 
        Less than 10 & 1,662 (26.67\%) & 0 & 5,660 (90.81\%) & 0 & 5,660 (90.81\%) \\ 
        10--100 & 3,672 (58.91\%) & 1.0--1.9 & 1 (0.02\%) & 1--9 & 231 (3.71\%) \\ 
        200--900 & 541 (8.68\%) & 2.0--2.9 & 1 (0.02\%) & 10--100 & 284 (4.56\%) \\ 
        1,000--5,000 & 237 (3.80\%) & 3.0--3.9 & 35 (0.56\%) & 200--1,000 & 55 (0.88\%) \\   
        10,000--50,000 & 106 (1.70\%) & 4.0--4.9 & 491 (7.88\%) & 5,000--100,000 & 3 (0.05\%) \\ 
        100,000--4,000,000 & 15 (0.24\%) & 5.0 & 45 (0.72\%) &  &  \\ \hline
        \bottomrule
    \end{tabular}}
\end{table}

\subsubsection{Capabilities}
Table~\ref{tab:gpt_capabilities} shows a wide variation in medical GPT capabilities, raising safety and privacy concerns in clinical contexts. While all models support basic conversation, many include advanced tools. Web Search is most common (71.15\%), enabling real-time external queries. Over half (53.63\%) support DALL.E image generation, 30.16\% support code interpretation for data analysis or computation, and 19.27\% support canvas, enabling collaborative writing and coding tasks. Actions that call external APIs account for just 2.73\% because they are more sensitive. GPT-4o's image tool appears in only 14.13\%, likely due to its recent launch. Nearly 25\% are text-only, while just 0.08\% support all six features. Most fall in the mid-range: 32.547\% support two tools, 29.41\% support three. This trend toward moderately equipped but increasingly connected models elevates privacy risks, especially when tools process sensitive health data. Therefore, there is a need for adequate safety analysis to examine both outputs and underlying technical configurations.

\subsubsection{Ethical Considerations}
We conducted this assessment in accordance with data ethics, platform policies, and institutional standards. All data were sourced from publicly available metadata on the OpenAI GPT Store, including GPT names, descriptions, conversation starters, categories, usage metrics, ratings, reviews, and capabilities. No user conversations, personal data, or private content were accessed. Engagement metrics (e.g., review and conversation counts) were treated as aggregate and non-identifiable, with no attempt to trace or link data to individuals. Actor-level abuse was evaluated using criteria derived from OpenAI's published policies, applied only to public GPT metadata and accessible privacy statements. We accessed no gated content, and data collection was performed via rate-limited queries in compliance with OpenAI's usage guidelines. All data are stored on institution-approved servers with access limited to authorized researchers.

\section{Measurement Methodology}\label{sec:methodology}
In this section, we introduce two frameworks to evaluate clinical hallucination and actor-intent misuse in MedGPTs across popularity tiers in the OpenAI Store.

To support large-scale interaction, we develop an automated tool using Selenium~\cite{selenium} to simulate user behavior through the web interface. This allowed us to prompt GPTs and collect responses as users would. Given OpenAI's account rate limits\footnote{Auditing all 6,233 models is practically and ethically infeasible under platform rate limits; our stratified design ensures representative coverage while following responsible Web auditing norms.}, we used both ChatGPT Plus and ChatGPT Pro subscriptions, allowing around 100 prompts every three hours. Testing all 6,233 MedGPTs was infeasible--it would take over a year. Instead, we sampled 1,500 models: the top 1,000 by conversation count, 250 from the middle tier, and 250 from the bottom tier, capturing behavior across the usage spectrum.

Following prior work~\cite{schmidgall2024addressingcognitivebiasmedical}, we restrict our interaction with LLMs -- both MedGPTs and open-source models -- to inference-only queries, excluding parameters such as temperature, gradients, or log-probabilities. This configuration mirrors the type of access available to typical users or patients, ensuring that our evaluation reflects realistic usage conditions.

\begin{table}[!t]
    \centering
    \footnotesize
\caption{Capabilities of MedGPTs.}
  \vspace{-0.2cm}
    \label{tab:gpt_capabilities}
   \tabcolsep=0.12cm
\renewcommand{\arraystretch}{1.2} 
\resizebox{0.48\textwidth}{!}{\begin{tabular}{ll||ll}
    \toprule
        \hline
        \textbf{\# Capabilities} & \textbf{\# GPTs} & \textbf{Capability} & \textbf{\# GPTs} \\ \hline 
        0 & 1,538 (24.68\%) & Web Search & 4,435 (71.15\%)  \\ 
        1 & 353 (5.66\%) & DALL.E Images & 3,343 (53.63\%) \\ 
        2 & 2,028 (32.54\%) & Code Interpreter \& Data Analysis & 1,880 (30.16\%) \\ 
        3 & 1,833 (29.41\%) & Canvas & 1,201 (19.27\%)  \\   
        4 & 408 (6.55\%) & 4o Image Generation & 881 (14.13\%) \\
        5 & 68 (1.09\%) & Actions & 170 (2.73\%)\\
        6 & 5 (0.08\%) &  & \\ \hline
        \textbf{Total} & \textbf{6,233} & & \\ \hline
        \bottomrule
    \end{tabular}}
    \vspace{-0.4cm}
\end{table}

\subsection{MedGPT-HEval: A Clinical Hallucination Evaluation Framework}\label{subsec:MedGPT-HEval}
We introduce MedGPT-HEval, a structured multi-metric framework for evaluating clinical hallucination in MedGPTs. We queried each GPT five times with diagnostic questions from the MedQA benchmark~\cite{MedQA}, framed within a clinical vignette to ensure contextual consistency. To capture variability in LLM outputs, we repeated the same question five times to obtain multiple responses and evaluated them using four complementary metrics: G-Eval~\cite{G-Eval}, BARTScore~\cite{BARTScore}, Semantic Entropy~\cite{Semantic_Entropy}, and Cosine Similarity~\cite{Cosine_similarity}.  

\subsubsection{G-Eval}
We use the G-Eval framework~\cite{G-Eval} to assess response faithfulness based on the clinical prompt and standardized context. Leveraging Gemini's chain-of-thought reasoning, each response is independently scored using a five-part rubric: (1) consistency with the prompt, (2) factual accuracy, (3) completeness, (4) citation reliability, and (5) inference justification. Scores (0–1 scale) are averaged across all responses. G-Eval uses Gemini 3.1 Pro as a judgment engine -- chosen for its advanced reasoning and intelligence -- to evaluate outputs through explicit steps: checking factual consistency, validating cited sources, and assessing logical coherence. We incorporate token-level confidence to mitigate bias from verbosity. The final score is the mean of the rubric components.

\subsubsection{BARTScore}
BARTScore~\cite{BARTScore} estimates the likelihood of a generated text ($y$) given a reference input ($x$) using the pretrained BART model~\cite{BioBART}. It assesses faithfulness by computing the weighted log-probability of $y$ conditioned on $x$, reflecting how closely the generated response aligns with the prompt’s meaning and structure. In our case, $x$ is the clinical vignette and question, and $y$ is the response generated by each MedGPT. Higher scores indicate better alignment and lower hallucination risk.

\subsubsection{Semantic Entropy}
Semantic Entropy (SE)~\cite{Semantic_Entropy} quantifies the model's uncertainty in generating outputs. Unlike prior formulations that require multiple samples and semantic-equivalence comparisons~\cite{Farquhar2024hallu}, we compute SE directly from a single forward pass using token-level output probabilities:

\[
SE(x) = - \sum_i p_i \log p_i
\]

where $p_i$ is the softmax probability of token $i$ for the generated sequence. This approach captures uncertainty and potential risk of hallucination. A higher $SE(x)$ indicates greater uncertainty in the model's predictions, which may signal a higher likelihood of hallucination.

\subsubsection{Cosine Similarity}
We use cosine similarity~\cite{Cosine_similarity} to assess semantic alignment between a medical GPT's response and its clinical input, defined as the concatenation of case context and question. Embeddings are generated with BioBERT, a transformer pretrained on biomedical text to ensure accurate encoding of clinical semantics. Scores range from 0 (no similarity) to 1 (perfect alignment). Lower scores indicate semantic drift or fabrication, key indicators of hallucination.

\subsection{Actor-level Abuse Evaluation Framework}\label{subsec:actor-level} 

\subsubsection{Method}
To evaluate design-intent risks, we developed an automated LLM-based scoring framework to assess MedGPT metadata. We focused on models with sufficient descriptive detail--specifically, those including at least two of name, description, or conversation starters--as these best reflect developer intent. Each model was evaluated for compliance with OpenAI's usage policies~\cite{OpenAI_privacy}, focusing on four misuse categories: health consultations, scams, privacy violations, and illicit activities.

A key challenge is the vagueness of OpenAI's policies. For instance, while ``providing medical advice'' is prohibited, the role of disclaimers or scope limitations remains unclear. To resolve this, we translated policy language into concrete criteria using LLM-as-a-judge methods~\cite{LLM-as-a-judge}, embedding these rules into structured Gemini 3.1 Pro prompts to ensure consistent and reproducible assessments. Gemini 3.1 Pro was chosen for its newness, strong smartness, advanced reasoning and intelligence, capability to solve complex problems, and low hallucination rate~\cite{gemini_2026}.

Each prompt included the GPT's metadata, a relevant policy clause, and a structured scoring rubric \footnote{\url{https://anonymous.4open.science/r/medical_llms-appendix-10DA}}. Gemini 3.1 Pro returned a misuse risk score and, when applicable, the specific policy or provision violated.

\subsubsection{Determining Threshold}\label{subsubsec:actor-level-threshold} 
To separate non-compliant MedGPTs from compliant ones, we applied K-means clustering with $k=2$ to the misuse risk scores. This unsupervised approach avoids reliance on arbitrary cutoffs and aligns with prior work~\cite{GPTRACKER} that uses similar methods to flag unsafe GPT behavior. The clustering produced a decision boundary at 0.45, which we adopt as the threshold for flagging non-compliant behavior. To validate this threshold, we conducted an expert review on models near the boundary (misuse scores 0.35-0.55). Expert judgments were compared with the clustering results. This yielded a Cohen's Kappa score of 0.8188, indicating almost perfect agreement. The silhouette score for the clustering was 0.7555, indicating that the two clusters are well separated and cohesive. These results confirm that the 0.45 threshold reliably separates non-compliant and compliant MedGPTs. 

\section{Results}
In this section, we present the empirical findings of our study, demonstrating how clinical hallucination and actor-level misuse manifest in MedGPTs across different popularity tiers. Afterwards, we show the pervasiveness of clinical hallucination in open-source medical LLMs and a comparative analysis with MedGPTs.

\subsection{Clinical Hallucination in MedGPTs}\label{subsec:results_hallucination}

\begin{figure*}[!t]
    \centering
    \vspace{-0.2cm}
    \subfigure[G-Eval]{%
        \includegraphics[width=0.2384\linewidth]{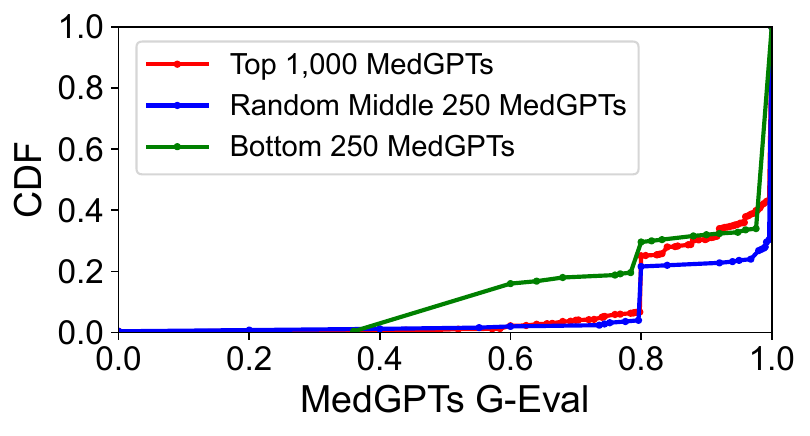}
        \label{fig:hallu_g-eval}
    }
    \subfigure[BARTScore]{%
        \includegraphics[width=0.2384\linewidth]{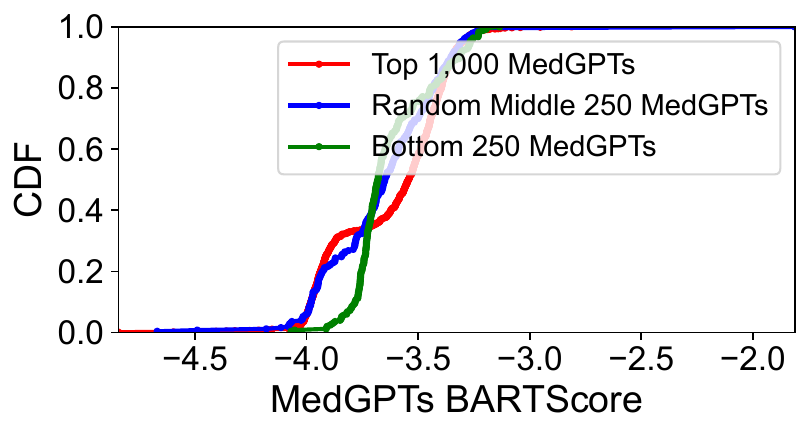}
        \label{fig:hallu_bartscore}
     }
    \subfigure[Semantic Entropy]{%
        \includegraphics[width=0.2384\linewidth]{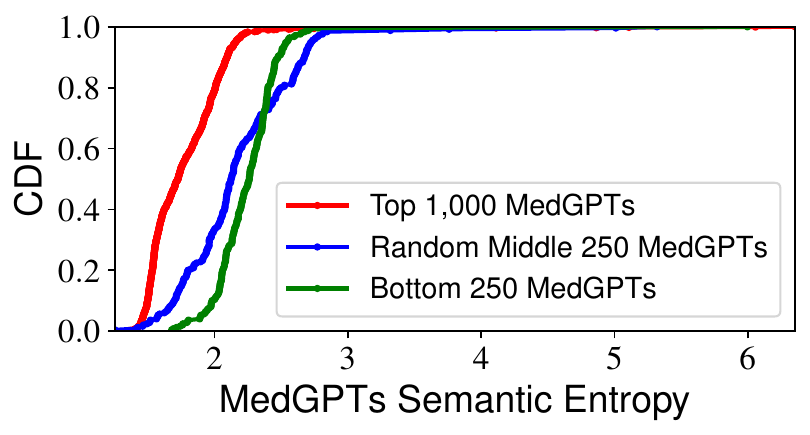}
        \label{fig:hallu_semantic}
    }
    \subfigure[Cosine Similarity]{%
        \includegraphics[width=0.2384\linewidth]{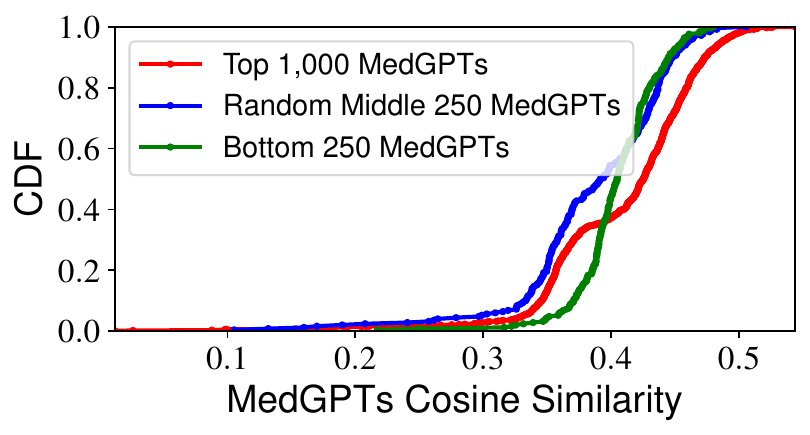}
        \label{fig:hallu_cosine}
    }
    \vspace{-0.2cm}
    \caption{Cumulative distribution of MedGPTs hallucination scores.}
    \label{fig:hallucination_scores}
    \vspace{-0.5cm}
\end{figure*}

Fig.~\ref{fig:hallucination_scores} shows the distribution of hallucination scores across MedGPTs grouped by popularity tier. As illustrated in Fig.~\ref{fig:hallu_g-eval}, 25\%, 22\%, and 30\% of MedGPTs in the top-, middle-, and bottom-tier categories score below 0.8, respectively. Similarly, 30\%, 22\%, and 32\% of models in the top, middle, and bottom tiers score below 0.9. Consequently, 70\%, 78\%, and 68\% of MedGPTs in these tiers achieve G-Eval scores of at least 0.9. Across all evaluated MedGPTs, 25.13\% score 0.8 or lower, indicating a non-trivial level of hallucination risk in the ecosystem. We also identify four MedGPTs in the top tier and one in the middle tier with a G-Eval score of zero. Our manual inspection shows that these models refused to respond to the diagnostic prompt, stating that they could not provide medical advice. Although such refusals reduce hallucination risk, they also reveal inconsistent safety or capability boundaries among deployed MedGPTs. Overall, bottom-tier MedGPTs exhibit the highest hallucination risk, followed by top-tier models, while mid-tier systems demonstrate comparatively lower hallucination rates. These results suggest that popularity does not necessarily correlate with factual reliability, raising concerns about the trust users may place in widely used web-deployed medical GPT applications.

Fig.~\ref{fig:hallu_bartscore} presents the BARTScore distribution for MedGPTs across the three categories. Approximately 5.9\%, 5.6\%, and 0.2\% of MedGPTs in the top, middle, and bottom categories, respectively, score below -4, indicating weak alignment between the generated responses and the reference context. The majority of MedGPTs fall within the -4 to -3.5 range, accounting for 52\%, 64.4\%, and 74.4\% of models in the top, middle, and bottom categories, respectively. Only 42.1\%, 30\%, and 24.8\% of MedGPTs in the top, middle, and bottom categories achieve scores of -3.5 or higher, suggesting stronger semantic alignment with the reference content. Overall, only 37.27\% of MedGPTs reach this threshold. These results indicate that many MedGPTs generate responses with limited semantic alignment to the clinical context, including several top-rated ones. This questions the reliability of MedGPTs for context-grounded clinical information and reveals the risk of inaccurate outputs that could influence medical guidance.

Fig.~\ref{fig:hallu_semantic} reports the semantic entropy scores across the three popularity categories. About 21\%, 66.8\%, and 90\% of MedGPTs in the top, middle, and bottom categories, respectively, record entropy scores $\ge$ 2.5. In contrast, 79\%, 33.3\%, and 10\% achieve scores below 2. Because semantic entropy is a cost metric -- where lower values indicate more stable, consistent outputs -- top-tier MedGPTs exhibit substantially lower uncertainty than middle- and bottom-tier MedGPTs. Overall, 59.87\% of MedGPTs obtain entropy scores below 2, suggesting moderate response stability across the ecosystem but clear performance gaps between popularity tiers. The middle- and bottom-tier MedGPTs show greater variability in generated responses, which may lead to inconsistent or unreliable clinical content in real-world healthcare applications.

Fig.~\ref{fig:hallu_cosine} presents cosine similarity across the three popularity categories and shows clear tier differences. The middle category has the largest share of low-alignment models (54.4\% $< 0.4$), compared with 43.6\% in the bottom category and 37.1\% in the top category. The bottom category is concentrated most heavily in the 0.4–0.45 band (48.8\%), exceeding the top (35.1\%) and middle (36\%) categories. Models with strong alignment ($> 0.45$) appear far more often in the top category (27.8\%) than in the middle (9.68\%) or bottom (7.6\%). Overall, 41.07\% of MedGPTs score below 0.4. Lower cosine similarity outside the top tier suggests that many models provide responses that do not align with the given context, increasing the risk of irrelevant or misleading content.

\noindent\textbf{Human annotation:} To validate our automated scoring, we manually reviewed a stratified subset of 300 MedGPTs: 200 from the Top 1000 tier and 50 each from the middle and bottom tiers ($\approx$20\% per tier) -- given the large number of models and the impracticality of reviewing all samples. Each response was independently assessed against the ground truth using the same metrics as the automated system, and discrepancies were resolved by consensus. Human reviewers agreed with the automated scoring in 90.5--94.0\% of cases (Cohen's kappa = 0.8163), confirming the robustness and reliability of our automated evaluation framework.

\subsection{Users' Perception of Clinical Hallucination}\label{subsec:results_users_awareness}
\begin{figure*}[!t]
    \centering
    \vspace{-0.2cm}
    \subfigure[G-Eval]{%
        \includegraphics[width=0.238\linewidth]{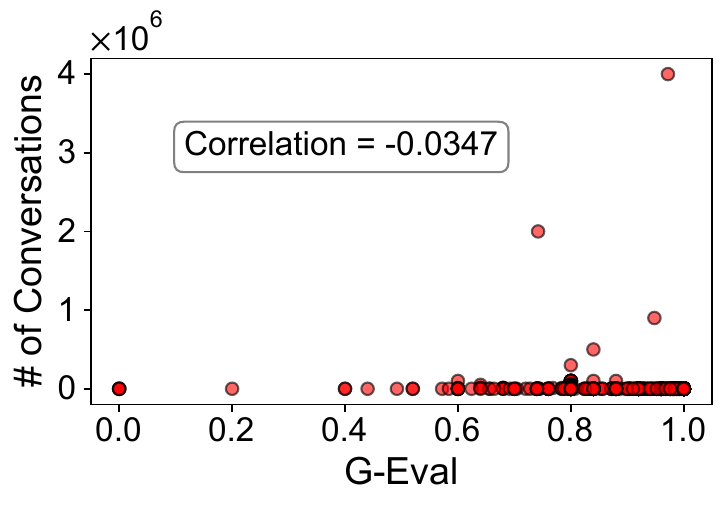}
        \label{fig:hallu_vs_conversation_top_geval}
    }
    \subfigure[BARTScore]{%
        \includegraphics[width=0.238\linewidth]{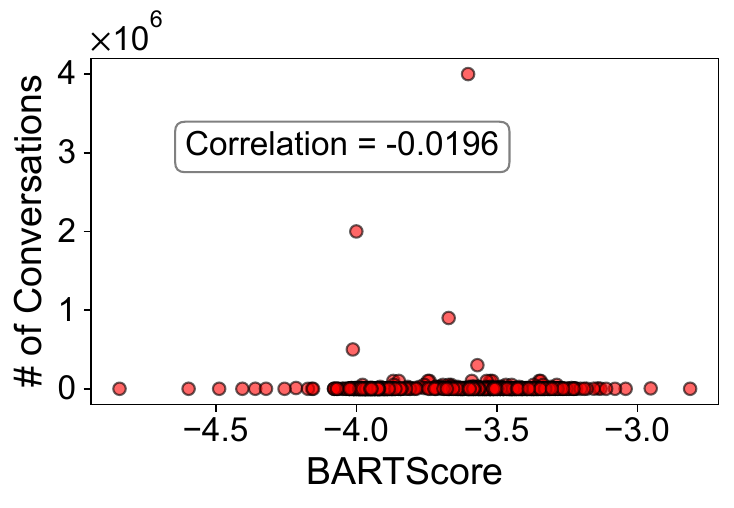}
        \label{fig:hallu_vs_conversation_top_bartscore}
     }
    \subfigure[Semantic Entropy]{%
        \includegraphics[width=0.238\linewidth]{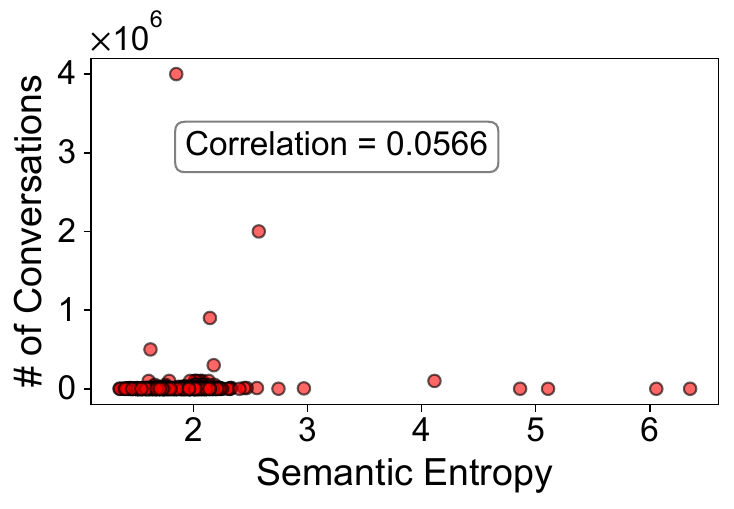}
        \label{fig:hallu_vs_conversation_top_semantic}
    }
    \subfigure[Cosine Similarity]{%
        \includegraphics[width=0.238\linewidth]{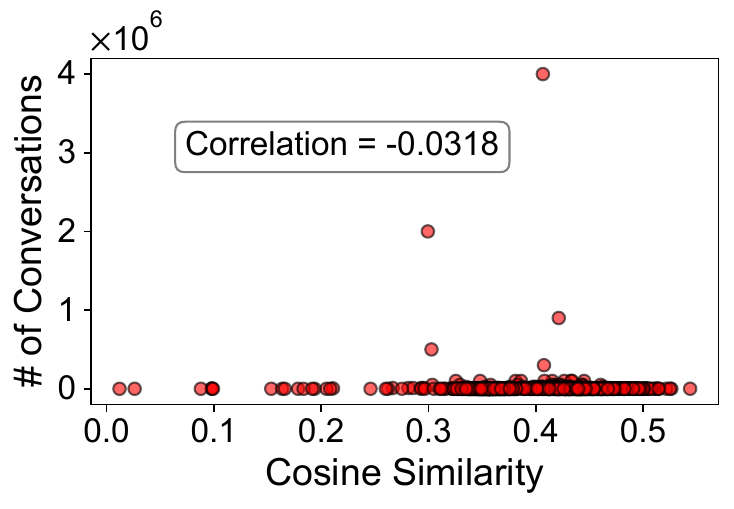}
        \label{fig:hallu_vs_conversation_top_cosine}
    }
    \vspace{-0.2cm}
    \caption{Correlation between hallucination scores and conversation counts of Top 1000 MedGPTs.}
    \label{fig:hallucination_vs_conversation}
\end{figure*}

\begin{figure}[h]
  \centering
  \vspace{-0.4cm}
  \subfigure[Positive reviews]{%
    \includegraphics[width=0.47\linewidth]{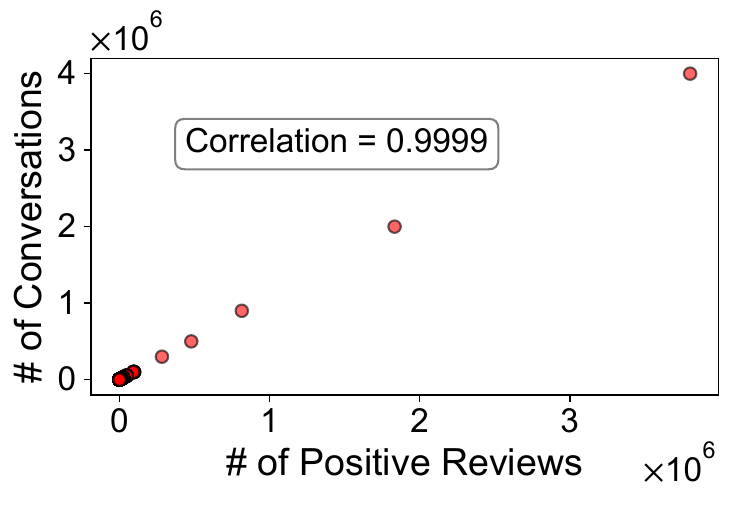}
    \label{fig:pos_review_hallu}
  }\hfill
  \subfigure[Negative reviews]{%
    \includegraphics[width=0.47\linewidth]{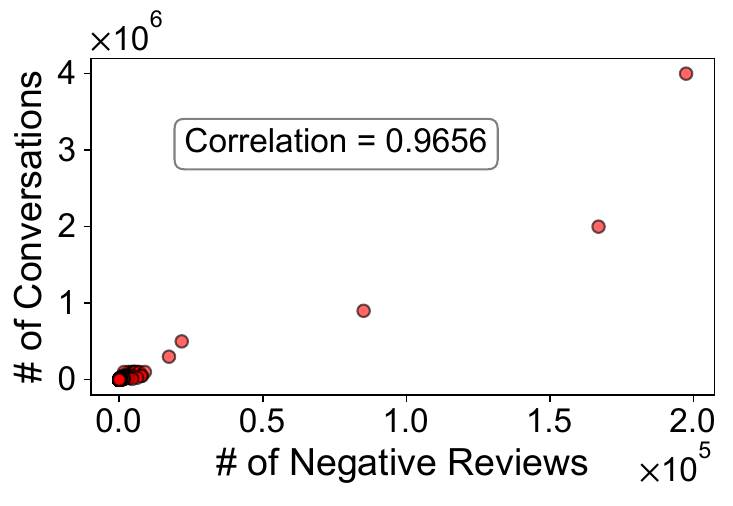}
    \label{fig:neg_review_hallu}
  }
  \vspace{-0.2cm}
  \caption{Correlation between conversation counts and reviews of Top 1000 MedGPTs.} 
  \label{fig:conversation+reviews}
  \vspace{-0.5cm}
\end{figure}

We evaluated whether user engagement reflects awareness of clinical hallucinations in MedGPTs. Fig.~\ref{fig:hallucination_vs_conversation} shows correlations between hallucination metrics and conversation counts for the Top 1000 MedGPTs, the only tier with sufficient review data. The four hallucination metrics are largely decoupled from user activity, with rank correlations of -0.0347, -0.0196, 0.0057, and -0.0318 for G-Eval, BARTScore, semantic entropy, and cosine similarity, respectively, suggesting that users rarely recognize or penalize clinical errors. Table~\ref{tab:hallucination_vs_review} confirms that hallucination scores are uncorrelated with both positive and negative reviews (ranging from -0.0449 to 0.0732), whereas reviews correlate strongly with usage ($r = 0.9999$ positive, $r = 0.9656$ negative), as shown in Fig.~\ref{fig:conversation+reviews}. These results reveal that user feedback reflects activity rather than accuracy, highlighting a critical gap between perceived and actual reliability in healthcare applications.

\begin{table}[h]
    \centering
\tabcolsep = 0.1cm
    \caption{Correlation between hallucination scores and reviews of Top 1000 MedGPTs.}
    \vspace{-0.2cm}
    \begin{tabular}{lllll}
        \toprule
        Review & G-Eval$\uparrow$ & BARTSCore$\uparrow$ & Semantic Entropy$\downarrow$ & Cosine Similarity$\uparrow$ \\
        \midrule
        Positive & -0.0341 & -0.0192 & 0.0555 & -0.0313 \\
        Negative & -0.0449 & -0.0260 & 0.0732 & -0.0388 \\
        \hline
    \end{tabular}
     \label{tab:hallucination_vs_review}
\end{table}

\subsection{Actor-defined Design Intent}\label{subsec:results_actor-defined}
\begin{figure}[!t]
    \centering
    \includegraphics[width=0.7\linewidth]{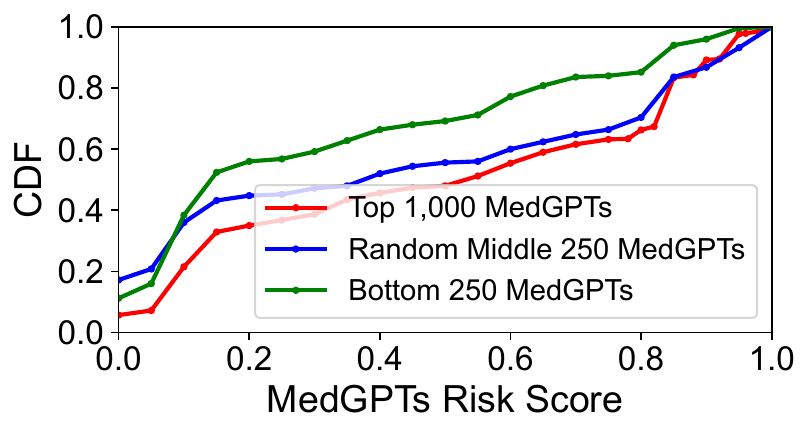}
     \vspace{-0.2cm}
    \caption{Cumulative distribution of misuse in MedGPTs.}
    \label{fig:risk_score_CDF}
\end{figure}

\begin{figure}[!t]
  \centering
  \subfigure[]{%
    \includegraphics[width=0.8\linewidth]{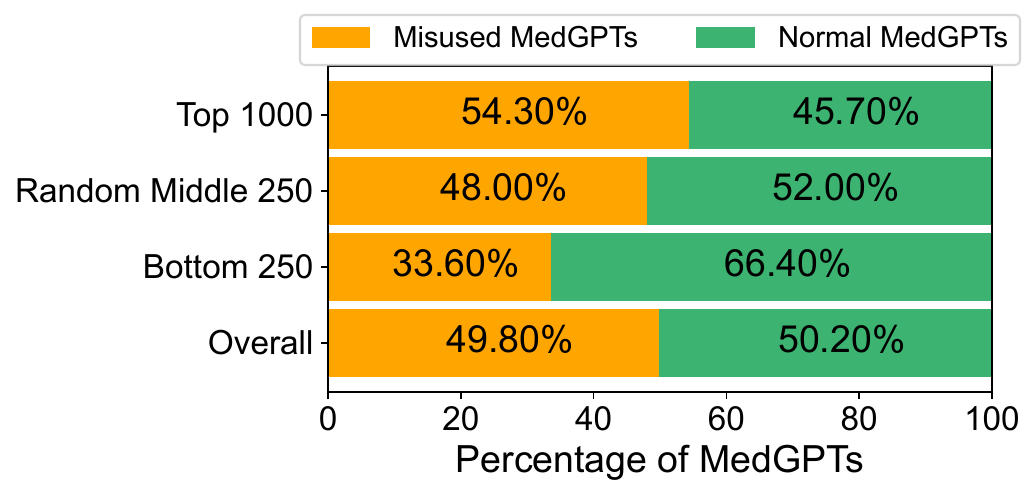}%
    \label{fig:misuse_stacked}
  }
  \subfigure[]{%
    \includegraphics[width=0.8\linewidth]{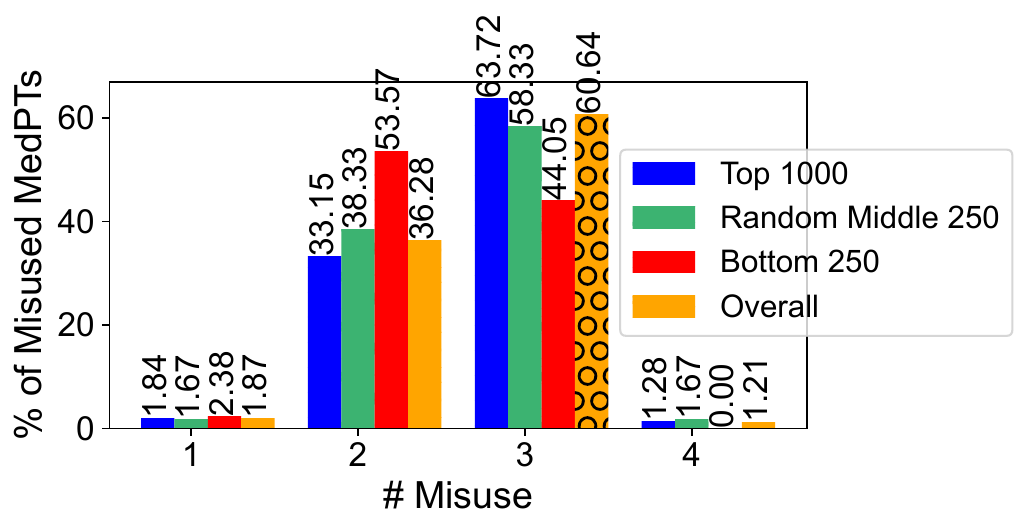}%
    \label{fig:misuse_bar}
  }
  \vspace{-0.3cm}
  \caption{Statistical distribution of misused MedGPTs.} 
  \label{fig:statistical_misuse}
\end{figure}

To assess risks beyond clinical hallucinations, we analyzed whether developer-defined design intent contributes to abusive behavior in MedGPTs. Fig.~\ref{fig:risk_score_CDF} shows the distribution of risk scores across popularity tiers. While the Top 1000 MedGPTs exhibit slightly lower risk than the middle and bottom tiers, abuse is present across all categories. Using the 0.45 threshold (Section~\ref{subsubsec:actor-level-threshold}), we identify misuse in 54.3\% of Top 1000, 48.0\% of Middle 250, and 33.6\% of Bottom 250, resulting in an overall misuse rate of 49.8\% (Fig.~\ref{fig:misuse_stacked}). These findings demonstrate that misuse is a systemic issue stemming from how developers encode design intent. Fig.~\ref{fig:misuse_bar} further breaks down abuse types. It shows that most violations involve two or three cases, affecting 33.15\% to 63.72\% of MedGPTs across tiers. Two-violation cases are most common in the bottom tier, while three-violation cases dominate the top and middle tiers. A small fraction of MedGPTs violate one case (Top: 1.84\%, Middle: 1.67\%, Bottom: 2.38\%) or four cases (Top: 1.28\%, Middle: 1.67\%). Although isolated violations are common, cumulative violations amplify the risk, especially in popular MedGPTs.\\ 

\begin{table*}[h]
\footnotesize
 \caption{Guidelines for disagreed cases.}
 \vspace{-0.2cm}
    \centering
    \tabcolsep = 0.12cm
    \begin{tabular}{llp{13.5cm}}
    \toprule
        Case  & \# & Description\\
        \midrule
        No conversation starters & 7 & These MedGPTs lack conversation starters, and neither their names nor descriptions indicate any policy violations \\
        No significant description  & 4 & This MedGPT lacks a meaningful description, and neither its name nor conversation starters suggest any violation\\
        No visible evidence & 16 & These MedGPTs include names, descriptions, and conversation starters, yet none indicate any violation\\
        The use of ``more'' in description & 3 & This MedGPT contains a name, description, and conversation starters, with no indication of violation other than the ambiguous use of ``more'' in the final part of its description \\
        \bottomrule
    \end{tabular}
    \label{tab:diagreement}
\end{table*}

\noindent\textbf{Human annotation:} To validate our automated scoring system and confirm that misused MedGPTs violated OpenAI's privacy policies, we manually reviewed a random 20\% of models using the same procedure described in Section~\ref{subsec:results_hallucination}. Each model’s name, description, and conversation starters were evaluated according to the policy taxonomy in Table~\ref{tab:misuse_description}. Human assessments agreed with the automated scoring in 89.5–92.0\% of cases (Cohen's kappa = 0.9479). Most discrepancies, detailed in Table~\ref{tab:diagreement}, arose from vague descriptions, missing conversation starters, or ambiguous phrasing, demonstrating both the reliability of our system and its limitations in instances with sparse metadata.

\subsection{MedGPTs with Actions capability}\label{subsec:results_actions_capability}
We analyzed MedGPTs with Actions capability to assess whether their published privacy policies reduce user risk or reflect non-compliance. Of the 170 MedGPTs with Actions enabled, we identified 108 unique third-party domains. As shown in Table~\ref{tab:policy_accessibility_distribution}, only 73 (42.94\%) had accessible privacy policies. Seven MedGPTs are linked to an unrelated policy. Most (66, 38.82\%) listed a single domain with one accessible policy, while four listed two domains, two listed three, and one listed four, often reusing the same policy. These inconsistencies suggest that while some creators follow disclosure standards, many offer minimal or mismatched documentation, undermining transparency.

The remaining 97 MedGPTs (57.06\%) lacked accessible policies. Of these, 62 (36.47\%) disclosed no third-party domains. In contrast, 35 listed domains with nonfunctional links: 13 were invalid, 8 redirected to generic homepages (including one OpenAI Enterprise instance), 11 led to suspended services, and 3 failed due to a DNS error (Table~\ref{tab:policy_accessibility_distribution}). These failures prevent users from understanding how their data is collected or shared. In clinical contexts, where inputs may include sensitive health information, the absence of clear policies undermines consent, compliance, and trust. 

\begin{figure}[!t]
  \centering
  \subfigure[]{%
    \includegraphics[width=0.45\linewidth]{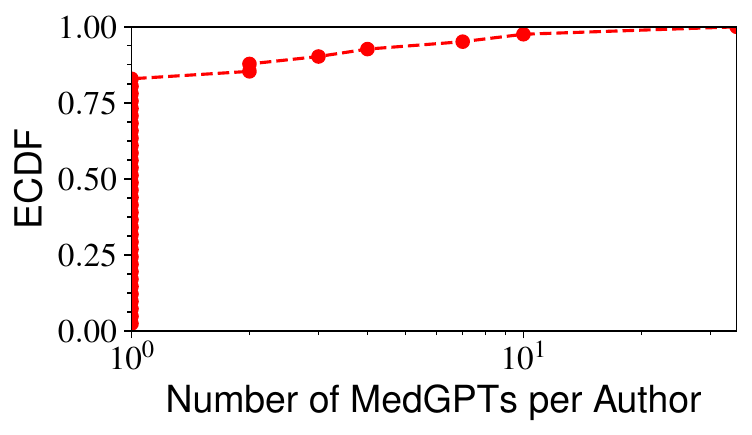}%
    \label{fig:author_actions_noncompliant}
  }\hfill
  \subfigure[]{%
    \includegraphics[width=0.45\linewidth]{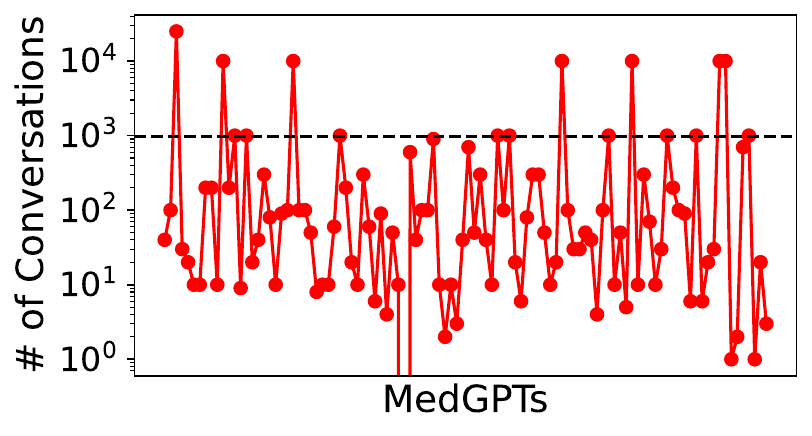}%
    \label{fig:conversation_actions_noncompliant}
  }
  \vspace{-0.3cm}
  \caption{Distribution of authors and conversation counts for noncompliant MedGPTs.}
  \vspace{-0.2cm}
  \label{fig:noncompliant}
\end{figure}

Noncompliant GPT authorship is highly concentrated (Fig.~\ref{fig:author_actions_noncompliant}). While $\approx$80\% were single-authored, two authors alone account for 35 and 10 models, respectively. These are not obscure: conversation counts range from 0 to over 25,000, averaging 984 (Fig.~\ref{fig:conversation_actions_noncompliant}). This indicates widespread use despite poor privacy protections.

\begin{figure}[!t]
    \centering
    \vspace{-0.2cm}
    \includegraphics[width=0.52\linewidth]{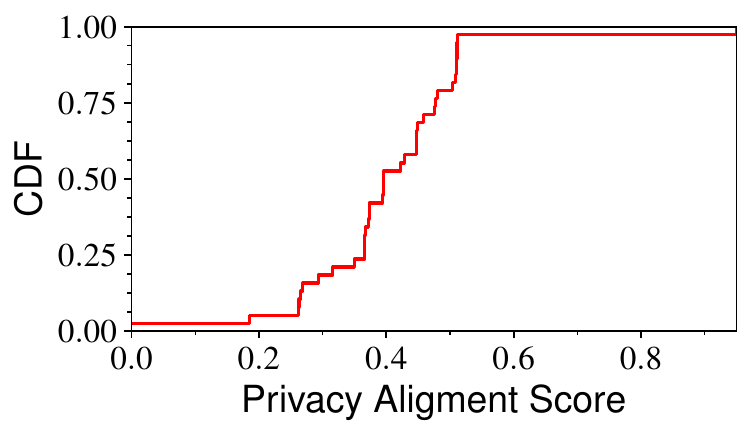}%
    \caption{Distribution of privacy policy alignment scores.}
    \label{fig:author_policy_cdf}
    \vspace{-0.3cm}
\end{figure}

Our automated analysis of 38 extracted privacy policies reveals major compliance gaps. As shown in Fig.~\ref{fig:author_policy_cdf}, nearly 70\% scored below 0.45, despite high Gemini 3.1 Pro confidence scores (0.713–0.968). Common lapses include missing disclosures on data sharing, cookies, analytics, user rights, and legal processing bases. These lapses reflect widespread neglect of privacy standards, increasing regulatory exposure and embedding unsafe data practices into clinical workflows.

\noindent \textbf{Human Review:} To validate the automated assessment, we manually reviewed these 38 policy statements and compared them with human labels. This produces a Cohen's Kappa of 0.8143, indicating almost perfect agreement. Thresholds for non-compliance were calibrated based on this validation, ensuring robust detection of privacy risks across Actions-enabled MedGPTs.

\begin{table}[!t]
  \centering
  \small
  \caption{Distribution of privacy policy accessibility.}
      \vspace{-0.2cm}
  \label{tab:policy_accessibility_distribution}
  \setlength{\tabcolsep}{0.12cm}
  \renewcommand{\arraystretch}{1.2}
  \resizebox{0.48\textwidth}{!}{%
  \begin{tabular}{c|c|cccc}
    \hline
    \multicolumn{1}{c|}{\textbf{Without domain}} & \multicolumn{5}{c}{\textbf{With domain}} \\
     \hline
    \multirow{4}{*}{62} & \multicolumn{5}{c}{108} \\
    \cline{2-6}
    \multicolumn{1}{c|}{} & \textbf{Accessible} & \multicolumn{4}{c}{\textbf{Inaccessible}} \\
    \cline{2-6}
     & \multirow{2}{*}{73} & \textbf{Broken link} & \textbf{Disabled service} & \textbf{Official website} & \textbf{DNS failure} \\
    \cline{3-6}
     &  & 13 & 11 & 8 & 3 \\
    \hline
  \end{tabular}%
  }
\end{table}

\subsection{Clinical Hallucination in Open-source Medical LLMs}\label{subsec: Hallucination_in_open-ssource LLMs}
We evaluate clinical hallucination in open-source medical LLMs using the methodology described in Section~\ref{sec:methodology}, with results summarized in Table~\ref{tab:open-source_model_hallu}. Galactica and Aloe-Alpha achieve the highest G-Eval scores (0.6480 and 0.5948), indicating stronger factual accuracy, whereas BioMistral and ChatDoctor score the lowest, suggesting a higher risk of hallucination. Cosine similarity highlights semantic consistency: MedAlpaca (0.4863), Apollo (0.4863), and MentalHealthChatbot (0.4354) outperform Aloe-Alpha (0.2298), revealing that high factual accuracy does not always align with coherent responses. Entropy values further capture response uncertainty, with MentalHealthChatbot (1.2077) and Galactica (1.2345) producing more stable outputs, while BioMistral (2.3978) is the most variable. BART scores reinforce these distinctions, with MedAlpaca and Apollo (-1.6549) exhibiting comparatively lower hallucination tendencies. Overall, the results reveal a trade-off between factual accuracy and response consistency, emphasizing the need for multi-metric evaluation when assessing hallucination in medical LLMs.

We compare MedGPTs with open-source medical LLMs using four metrics, as shown in Table~\ref{tab:MedGPT_vs_Open-source}. MedGPTs achieve significantly higher average G-Eval (0.9238 vs. 0.4558), indicating stronger factual accuracy, though their performance ranges from 0 to 1, revealing occasional low-quality outputs. BART scores show that open-source models maintain slightly higher average (-3.5310 vs. -3.6307) and maximum (-1.6540 vs. -1.8115) values, indicating more reliable outputs. Semantic entropy is lower for open-source models (1.6129 vs. 1.9272), suggesting that their outputs are more consistent and predictable. Cosine similarity is higher on average for MedGPTs (0.4054 vs. 0.3731), reflecting better semantic alignment. These results demonstrate that MedGPTs provide superior factual accuracy and semantic coherence, while open-source medical LLMs remain more consistent and reliable on certain metrics. These findings demonstrate the need to consider multiple evaluation dimensions when assessing hallucination in medical LLMs.

\begin{table}[!t]
\small
\caption{Open-source LLMs evaluation metrics.}
\vspace{-0.2cm}
\tabcolsep = 0.13cm
    \centering
    \scalebox{0.90} {\begin{tabular}{lllll}
    \toprule
        \textbf{Model} & \textbf{G-Eval$\uparrow$} & \textbf{BART$\uparrow$} & \textbf{Entropy$\downarrow$} & \textbf{Cosine$\uparrow$} \\
        \midrule
        Galactica~\cite{galactica} & 0.6480 & -3.8400  & 1.2345 & 0.3814 \\
        ChatDoctor~\cite{chatdoctor} & 0.2000   & -4.0408 & 1.9006 & 0.3573 \\
        MedAlpaca~\cite{medalpaca} & 0.3527 & -1.6549 & 1.7856 & 0.4863 \\
        PMC-LLaMA~\cite{pmc-llama} & 0.4724 & -4.1148 & 1.4092 & 0.3704 \\
        JMLR~\cite{jmlr} & 0.4202 & -4.0731 & 1.2931 & 0.3248 \\
        BioMistral~\cite{bioMistral} & 0.3869   & -4.0547 & 2.3978 & 0.296 \\
        Apollo~\cite{apollo} & 0.5060 & -1.6549 & 1.7856 & 0.4863 \\
        Aloe-Alpha~\cite{aloe-Aplha} & 0.5948 & -4.0824 & 1.6217 & 0.2298 \\
        MentalHealthChatbot~\cite{mentalHealthChatbot} & 0.4810  & -3.8392 & 1.2077 & 0.4354 \\
        Zhongjing~\cite{Zhongjing} & 0.4963 & -3.955 & 1.4932 & 0.3632 \\
        \bottomrule
    \end{tabular}
    }
    \label{tab:open-source model_hallu}
\end{table}

\begin{table}[h]
    \centering
    \tabcolsep = 0.12cm
    \caption{MedGPTs' Metrics versus Open-source LLMs.}
    \vspace{-0.2cm}
    \scalebox{0.90} {\begin{tabular}{lccc|ccc}
        \hline
       \multirow{2}{*}{Metric} & \multicolumn{3}{c|}{MedGPTs} & \multicolumn{3}{c}{Open-source LLMs} \\
        \cline{2-7}
        & Min & Max & Avg & Min & Max & Avg \\
        \hline
        G-Eval$\uparrow$ & 0 & 1 & 0.9238 & 0.2000 & 0.6480 & 0.4558 \\
        BARTScore$\uparrow$ & -4.8428 & -1.8115 & -3.6307 & -4.1148 & -1.6540 & -3.5310 \\
        Semantic Entropy$\downarrow$ & 1.2535 & 6.3543 & 1.9272 & 1.2077 & 2.3978 & 1.6129 \\
        Cosine Similarity$\uparrow$ & 0.0121 & 0.5438 & 0.4054 & 0.2298 & 0.4863 & 0.3731 \\
        \hline
    \end{tabular}
    }
     \label{tab:MedGPT_vs_Open-source}
     \vspace{-0.3cm}
\end{table}

\section{Discussion and Recommendations}
\subsection{Discussion}
This subsection outlines the broader implications of our findings and proposes concrete steps to improve safety, accountability, and patient trust in medical LLMs.

\textit{Hallucination is systemic, not incidental.} MedGPT-HEval reveals that 25--30\% of models across all tiers score below 0.8 in G-Eval, with fewer than 42\% achieving adequate BARTScore or cosine similarity, and only $\approx$60\% producing stable outputs (semantic entropy $<$ 2). Bottom- and middle-tier models pose the greatest risk, demonstrating that popularity is a poor predictor of factual reliability. Without multi-metric evaluation before deployment, hallucination remains a pervasive safety concern in web-deployed MedGPTs.

\textit{User unawareness of clinical hallucination.} Engagement data reveal a structural failure -- users consistently reward usage, not accuracy. While conversation counts and reviews are tightly correlated ($r \approx 1$), their correlation with clinical hallucination is negligible ($r < 0.06$), indicating that users rarely recognize clinical errors. In fact, less reliable MedGPTs often received more positive feedback -- suggesting that fluency is mistaken for trustworthiness. This mismatch allows unsafe models to gain prominence even when they offer misleading guidance. To address this, platforms should integrate standardized safety assessments into model profiles, aligning trust signals with actual reliability.

\textit{Abuse by design and unsafe by intent.} Misuse in MedGPTs is often deliberate. Over 49\% violated OpenAI's usage policies, many explicitly promoting unsafe consultations, scams, or illicit advice in their metadata. Over 33\% combined multiple violations, amplifying risk. These were not borderline cases, as our manual review confirmed the intent. This points to a developer ecosystem where harmful behavior is embedded at design time. Without enforced red-teaming and stronger accountability, such models will persist and scale.

\textit{Privacy as collateral damage}. Privacy violations were widespread. More than half of MedGPTs with Actions capability lacked accessible policies; among those available, $\approx$70\% failed basic compliance checks. Many were generic templates without key disclosures on data sharing, tracking, or user rights. These noncompliant GPTs are often widely used, exposing users to unsafe data handling. Enforceable policy standards, routine audits, and model takedowns are urgently needed to safeguard patient privacy in LLM-driven healthcare.

\textit{Clinical hallucination varies across medical LLMs.} Open-source medical LLMs show a 25--65\% trade-off between factual accuracy and semantic coherence: some models produce more accurate outputs, while others are more consistent or aligned with context. Compared with MedGPTs, which generally achieve higher factual accuracy and semantic alignment (G-Eval $\approx$0.92, cosine $\approx$0.41), open-source models are more stable but may sacrifice correctness. These findings emphasize that popularity or perceived quality does not guarantee reliability, underscoring the need for multi-metric evaluation to ensure safe clinical outputs from medical LLMs.

Because the OpenAI Store operates as a general-purpose AI marketplace, our findings likely generalize to other Web-scale LLM deployment ecosystems (HuggingFace spaces, Poe bots, Replit agents), making our framework broadly applicable.

\subsection{Recommendations}
Here, we outline concrete safeguards to mitigate clinical hallucination and unsafe design in medical LLMs.

\begin{itemize}[left=0pt]

    \item \textit{Hallucinations} in medical LLMs -- confident but unfounded claims -- pose serious safety risks. Mitigation requires evidence grounding, model-level refinement, and human oversight. Retrieval-augmented generation (RAG) anchors biomedical outputs, while hallucination-aware fine-tuning and reinforcement learning with human feedback (RLHF) improve factuality~\cite{gumaan2025}. Real-time systems such as detection-and-repair~\cite{varshney2023} and RARR~\cite{gao2023-rarr} revise outputs based on retrieved evidence. Human-in-the-loop frameworks~\cite{ahmad2023} can flag low-confidence responses, escalate to clinicians, and guide model updates. These approaches shift hallucination control from reactive to proactive, trustworthy design.
      
    \item Our analysis showed that 49.8\% of MedGPTs encode abusive behaviors, including unsafe consultation and illicit advice. Addressing this requires shifting from broad policy language to enforceable, developer-focused safeguards. The OpenAI Platform should translate usage policies~\cite{rodriguez2025safer} into clear, auditable rules (as our framework demonstrates), automatically evaluate models at submission~\cite{GPTRACKER}, and flag high-risk cases for review. Routine assessments of high-risk GPTs~\cite{GPTRACKER} and transparency dashboards showing model intent, compliance, and known risks can deter misuse. Without such measures, harmful models continue to scale, embedding risk into the foundations of healthcare LLMs.
    
    \item MedGPTs with Actions capability pose serious privacy risks: 57.06\% lacked accessible policies, and $\approx$70\% of retrieved ones failed basic compliance. Creators often reuse vague templates or link to broken or irrelevant pages, despite many models logging thousands of interactions. The OpenAI Platform should use our evaluation framework to flag noncompliant models, enforce policy validation at submission, reject invalid disclosures, and show privacy scores~\cite{GPTRACKER}. In clinical contexts, missing or misleading policies undermine consent, erode trust, and risk regulatory violations. Automated filters should detect breaches at scale, supplemented by human review to catch nuanced issues, exploitation, or unsafe behavior that automated systems may miss.
\end{itemize}

\subsection{Concluding Remarks}
We conducted the first large-scale audit of MedGPTs on the OpenAI Store, analyzing 6,233 models via automated extraction, structured prompting, and multi-metric scoring. We also evaluated hallucination risks in 10 open-source medical LLMs for comparison. Our framework identifies two systemic risks: clinical hallucinations and actor-driven misuse (abusive intent, privacy violations, unsafe Actions). Results show 25--30\% of MedGPTs score below 0.8 in G-Eval, 37.27\% reach a BART score $\geq-3.5$, 41.07\% have cosine similarity $\geq0.4$, and 59.87\% have semantic entropy $<2$. Misuse affects 54.3\% (top), 48.0\% (middle), and 33.6\% (bottom) tiers, and only 42.94\% of 170 Action-enabled models had accessible privacy policies, nearly 70\% of which scored below the threshold. Compared with MedGPTs, open-source models have lower G-Eval (0.456 vs. 0.924) but more stable semantic entropy (1.613 vs. 1.927). User engagement shows near-zero correlation with hallucination metrics, highlighting gaps in accuracy, consistency, and perceived reliability. 

This study has several limitations. \textit{First}, it reflects a snapshot of the OpenAI GPT Store (January 20--22, 2026), and trends may change. \textit{Second}, although we extracted 6,233 MedGPTs, coverage may be incomplete, and open-source evaluations focused on prominent models. \textit{Third}, we considered only clinical hallucinations and actor-driven risks (consultation abuse, scams, privacy violations, illicit activity, and Actions); other vectors, such as adversarial prompts or data leakage, remain unexplored. \textit{Fourth}, intent analysis relied on static metadata; dynamic abuses via third-party APIs may evade detection. \textit{Fifth}, due to OpenAI query limits, we analyzed a stratified subset rather than the full dataset. While representative, broader coverage may yield different results. Future work should extend to continuous, scalable audits, dynamic red-teaming, and standardized safety benchmarks for trustworthy medical LLMs.

\balance
\bibliographystyle{IEEEtran}
\bibliography{references}

\FloatBarrier

\begin{table*}[h]
   \footnotesize
 \caption{Operationalized policies and context for forbidden scenarios.}
     \label{tab:misuse_description}
    \centering
    \begin{minipage}{1.0\textwidth}
    \begin{tabular}{p{2.5cm}p{6.1cm}p{9.0cm}}
        \toprule
        Proscribed case & Context & Policy description \\ \midrule
        \multirow{6}{*}{Health consultation} & Using these models to advise someone on possible diagnoses, recommend medications or dosages, interpret lab results, or suggest actions in response to acute medical symptoms or crises. & OpenAI's models are not specifically designed or approved to offer medical guidance. They should not be used for diagnosing health conditions or recommending treatments. These tools are not to be used for handling emergencies or life-threatening situations requiring urgent care. It is the developers' or creators' responsibility to ensure compliance with relevant medical regulations when utilizing OpenAI services in any healthcare-related application. \\ \hline
        
        \multirow{7}{*}{Medical/Economic scam} & Developing or deploying applications, chatbots, marketing tools, or automated scripts that leverage OpenAI’s models to generate persuasive messaging intended to sell or endorse unproven medical treatments, therapies, or health-related products, particularly when such usage is aimed at influencing consumer behavior for profit. & OpenAI strictly forbids the use of its models or GPT-based tools to promote misleading medical treatments or health-related products for commercial gain.\\ \hline
        
        \multirow{16}{*}{Privacy Infringement} & Deploying GPT-based systems in applications that involve high-stakes decision-making, surveillance, user profiling, or the unauthorized collection or manipulation of personal data, such as tools used for facial recognition, legal or medical consultation, financial planning, social engineering, or deceptive user engagement, without ensuring full compliance with applicable laws, ethical standards, user consent protocols, and age-appropriate protections. & Engaging in any activity that compromises the privacy, rights, or safety of others is strictly prohibited. This includes, but is not limited to: (a) Collecting, processing, inferring, or generating personal data without adhering to relevant legal obligations. (b) Employing biometric technologies, such as facial recognition, for identifying or profiling individuals. (c) Monitoring individuals without consent or enabling the use of spyware or surveillance tools. (d) Delivering personalized advice in sensitive areas such as healthcare, law, or finance without appropriate qualifications or disclaimers. (e) Making automated decisions that significantly impact an individual’s rights, freedoms, or well-being. (f) Producing or disseminating false, misleading, or deceptive content, including disinformation, misinformation, or manipulative digital engagement. (g) Impersonating people or organizations without explicit permission or legal authority. (h) Misleading users about the intentions, capabilities, or function of your GPT or model. (i) Creating models designed for or targeting users under the age of 13, or offering content inappropriate for minors.\\ \hline
        
        \multirow{7}{*}{Illicit Activity} & Using these models to support illegal activities such as providing unlicensed medical advice, generating counterfeit prescriptions, promoting the sale of unauthorized drugs or treatments, falsifying health records or insurance claims, or engaging in other unlawful practices across healthcare, pharmaceuticals, or related sectors like insurance and telemedicine. & The use of OpenAI’s models, tools, or services for any unlawful purpose is strictly prohibited.\\ \bottomrule
    \end{tabular}
    \end{minipage}
\end{table*}

\end{document}